\documentclass[letterpaper, 10 pt, conference]{ieeeconf}
\IEEEoverridecommandlockouts
\makeatletter
\def\@IEEEtablestring{table}
\long\def\@makecaption#1#2{%
\ifx\@captype\@IEEEtablestring%
\noindent\parbox[t]{\hsize}{\footnotesize\raggedright #1:~#2}\par\@IEEEtablecaptionsepspace%
\else
\@IEEEfigurecaptionsepspace%
\setbox\@tempboxa\hbox{\footnotesize #1.~~ #2}%
\ifdim \wd\@tempboxa >\hsize%
\setbox\@tempboxa\hbox{\footnotesize #1.~~ }%
\parbox[t]{\hsize}{\footnotesize \noindent\unhbox\@tempboxa#2}%
\else%
\ifcenterfigcaptions \hbox to\hsize{\footnotesize\hfil\box\@tempboxa\hfil}%
\else \hbox to\hsize{\footnotesize\box\@tempboxa\hfil}%
\fi\fi\fi}
\makeatother
\usepackage{graphicx}
\usepackage{amsmath,amssymb}
\usepackage{booktabs}
\usepackage{array}
\usepackage{xcolor}
\usepackage{textcomp}
\usepackage[caption=false,font=footnotesize]{subfig}
\usepackage{cuted}
\usepackage{capt-of}
\usepackage{flushend}
\usepackage{fancyhdr}
\fancypagestyle{ieeenotice}{%
  \fancyhf{}%
  \fancyfoot[C]{\resizebox{\textwidth}{!}{%
    This work has been submitted to the IEEE for possible publication.
    Copyright may be transferred without notice, after which this version may no longer be accessible.}}%
}

\graphicspath{{figures/}}

\usepackage{microtype}
\makeatletter
\g@addto@macro\normalsize{%
  \setlength\abovedisplayskip{4pt plus 2pt minus 2pt}%
  \setlength\belowdisplayskip{4pt plus 2pt minus 2pt}%
  \setlength\abovedisplayshortskip{2pt plus 1pt}%
  \setlength\belowdisplayshortskip{3pt plus 1pt minus 1pt}%
}
\let\@oldthebibliography\thebibliography
\renewcommand{\thebibliography}[1]{\@oldthebibliography{#1}%
  \setlength{\itemsep}{0pt plus 0.3pt}%
  \setlength{\parsep}{0pt}%
  \setlength{\topsep}{2pt}}
\makeatother

\usepackage[pdfauthor={Jinseok Lee, Dongho Yee, Minsung Kim, Younghoon Noh, Seonho Shim,
  Juahn Oh, Yechan Seo, Jiyul Lee, Seong Jeong, Hyuk Choi, Hyoun-Joong Kong},
 pdftitle={Drape-Compatible Tool-Tip Localization for Hand-Held Laparoscopic Instruments
  via UWB Carrier-Phase Ranging and Trocar-Constrained Geometry},
 pdfsubject={UWB carrier-phase ranging for surgical instrument tracking},
 pdfkeywords={Medical Robots and Systems; Localization; Surgical Robotics: Laparoscopy;
  ultra-wideband; carrier phase; remote center of motion},
 hidelinks]{hyperref}

\begin{document}

\title{\LARGE \bf Drape-Compatible Tool-Tip Localization for Hand-Held Laparoscopic Instruments via UWB Carrier-Phase Ranging and Trocar-Constrained Geometry
}

\author{Jinseok Lee$^{*,2,4}$, Dongho Yee$^{*,1,2,4,5}$, Minsung Kim$^{1,2,4}$, Younghoon Noh$^{2,4}$,\\
Seonho Shim$^{2,8}$, Juahn Oh$^{1,2,7}$, Yechan Seo$^{1,2,3}$, Jiyul Lee$^{1,2,3}$, Seong Jeong$^{1,2,3}$,\\
Hyuk Choi$^{1,2,3}$ and Hyoun-Joong Kong$^{\dagger,1,3,6}$%
\thanks{$^{*}$First authors. $^{\dagger}$Corresponding author.}%
\thanks{\scriptsize $^{1}$Department of Transdisciplinary Medicine, Seoul National University
Hospital. $^{2}$Rosota Inc., Seoul, Republic of Korea. $^{3}$Department of Medicine,
Seoul National University College of Medicine. $^{4}$Department of Mechanical
Engineering, Seoul National University. $^{5}$Department of Computer Science
and Engineering, Seoul National University. $^{6}$Institute of Convergence
Medicine with Innovative Technology, Seoul National University Hospital.
$^{7}$Eulji University College of Medicine. $^{8}$Department of Mechanical
Engineering, Chungang University.}%
}
\maketitle
\thispagestyle{ieeenotice}
\pagestyle{empty}

\begin{strip}
\centering
\begin{tabular}{@{}c@{\hspace{10pt}}c@{}}
\raisebox{0.02in}{\includegraphics[width=0.43\textwidth]{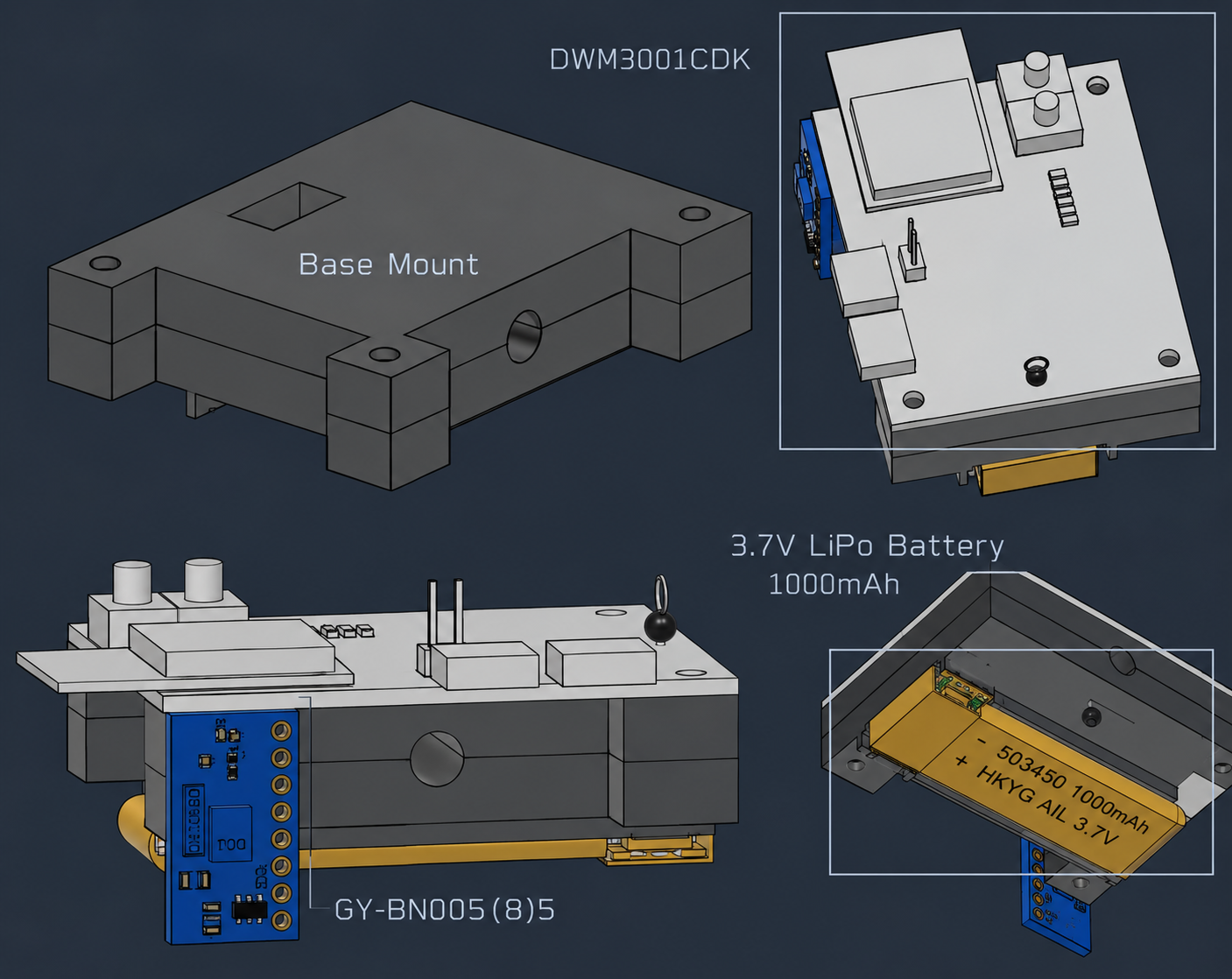}} &
\includegraphics[width=0.54\textwidth]{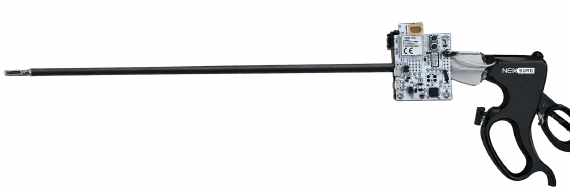}\\[3pt]
{\footnotesize (a) What one node is made of} &
{\footnotesize (b) The node clipped onto an unmodified instrument}
\end{tabular}
\captionof{figure}{\textbf{A clip-on node that ranges through the sterile drape.} Three of these
nodes, one near the handle of each hand-held instrument and one on the endoscope, sit outside the
drape and range to one another by carrier phase; the trocar then turns the three distances into the
two insertion depths from which the tool tips follow. (a)~A node, exploded: a 3D-printed base mount
that clips to the shaft, a Qorvo DWM3001CDK UWB board, a GY-BNO085 IMU on its side face, and a
1000\,mAh 3.7\,V LiPo cell in the mount, so the instrument carries no cable. (b)~The assembled node
on an unmodified 5\,mm laparoscopic instrument, between the handle and the trocar, with nothing
added to the part that enters the patient (Sec.~\ref{sec:platform},~\ref{sec:integrated}).}
\label{fig:teaser}
\end{strip}

\begin{abstract}
A surgical robot policy needs to know where each instrument's working end sits relative to the
camera and to the other instrument, yet a laparoscopic operation leaves only the endoscope video,
and a draped instrument hides every optical path on itself. We estimate the tool tips from
\emph{distances and an IMU alone}. The distances are measured by ultra-wideband carrier phase
between antenna nodes on the handle side of the instruments: one per hand-held instrument, one on
the endoscope, all outside the sterile barrier. Take the endoscope antenna as the reference. The
two instrument antennas carry six coordinates while the three pairs give three distances, so the
problem is short by three, and averaging cannot close that gap because what is missing is
information, not precision. Carrier phase adds one more
unknown per pair, an integer that leaves distance fixed only modulo $\lambda/2=23.1$\,mm. The
trocar settles both difficulties: each shaft passes through a port whose position is known and
whose direction the on-board IMU measures, so its antenna keeps a single degree of freedom, the
insertion depth. Two unknowns now face three distances --- two fix the depths and the third is left
over, and that spare equation exposes a wrong integer or a wrong mounting offset instead of
absorbing it. The same solve returns both tool tips in the endoscope frame, without the second
carrier frequency that integer resolution usually needs, so the link never leaves a single PLL
lock. In an 11-hole phantom among metal instruments the three distances hold
$\sigma=0.10$--$0.33$\,mm at 35--46\,Hz, a sterile drape pressed onto the antennas leaves the link
unchanged where an optical path would be blocked, and the logger was carried into a live rabbit
appendectomy.
\end{abstract}

\section{Introduction}\label{sec:intro}
Surgical robot policies are increasingly learned from human demonstrations, which outside surgery
come from hand-held recorders rather than from the robot itself. What such a
policy consumes is geometry at each timestep: where each instrument's working end sits relative to
the camera and to the other instrument. A conventional laparoscopic operation records none of it,
leaving only endoscope video in which the tips appear when nothing occludes them.

Optical trackers need line of sight to markers, while the part to be localized is inside the
patient; electromagnetic trackers degrade near the metal instruments that fill a surgical
field; marker-less endoscopic estimation degrades under occlusion. The operating room compounds all three: an instrument that enters it must be draped unless all its
electronics are sterilized, and the drape seals any optical path on the instrument inside
the barrier.

We therefore clip an ultra-wideband node near the handle of each hand-held instrument and one on
the endoscope, all outside the sterile barrier, and let them range against each other; the three
nodes are the \emph{kinematic logger}. Radio crosses the drape, and distance --- unlike a marker or
an image --- needs no line of sight. The exchange whose pulses give a coarse time of flight (ToF)
also carries the carrier's phase of arrival: at channel 5 one cycle spans 46.2\,mm, so a few
degrees of phase noise is tens of micrometers of path.

Keeping that phase continuous is the radio problem; turning three distances into two tool tips is
the geometric one. Three scalar distances do not locate two free-floating instruments, and phase fixes distance only
modulo $\lambda/2=23.1$\,mm, so a lost round or a motion larger than half a cell slips that
registration by one cell. The trocar addresses both: each instrument turns about a fixed port, so
its antenna lies on a line whose direction its on-board IMU measures and contributes one unknown,
the insertion depth. Two continuous unknowns now face three measured distances, and the estimate
becomes a small mixed integer--continuous least-squares problem whose solve returns both tool tips
in the endoscope frame. Unlike the standard treatment of carrier-phase ambiguity, which buys the
integer with more measurements, we buy it with kinematics the operating room already provides and
keep every exchange on one PLL lock --- our two-channel version slipped a cell every 30--60\,s
because each hop re-randomizes the phase.

\textbf{Contributions.}
\begin{itemize}
\item \textbf{Sub-mm carrier-phase ranging between hand-held instruments in a
multipath-rich field.} A \emph{post-final} packet that removes each round's residual
carrier-frequency term, crystal trimming, a per-round quality gate against multipath amplitude
nulls, and a single-channel three-node schedule together keep the phase chain continuous. Among metal instruments in the 11-hole phantom, the three inter-antenna
distances hold $\sigma=0.10$--$0.33$\,mm at 35--46\,Hz; on the rail, blind displacements of
25--53\,mm are tracked to 2.8\,mm mean absolute error against caliper-measured values, while ToF
from the same module mis-measures the same displacements by 13--149\,mm and scatters by
17--20\,mm shot to shot (Sec.~\ref{sec:radio},~\ref{sec:res-rail},~\ref{sec:res-static}).

\item \textbf{A pivot-constrained, range-only formulation that turns three distances into
two tool tips.} Each trocar confines its antenna to a line whose direction the IMU measures, so
three distances over-determine the two insertion depths by one. The same solve returns both tips in
the endoscope frame, and its residual exposes a large slip or a wrong mounting offset without a
second carrier frequency, while $\pm1$-cell steps are bounded by the Kalman gate. In synthetic
trials on the measured phantom geometry the $\pm5$-cell hypothesis set admits no false fix (0/200) and corrects 63\% of
injected slips, 82\% of its failures being silence rather than a wrong pick (Sec.~\ref{sec:judge},~\ref{sec:res-tip}).

\item \textbf{A drape-compatible kinematic logger.} Because a drape defeats vision markers and
optical ToF, all three nodes sit outside the sterile barrier and sense through it: with the drape
pressed onto both chip antennas the round rate is unchanged, first-path SNR stays within 7\% of
bare, and the raw static $\sigma$ does not move. The logger was then carried into a live rabbit
appendectomy, recording 11 episodes (133\,s) of three-pair ranging with the working ends inside
the animal
(Sec.~\ref{sec:res-rail},~\ref{sec:res-invivo}).
\end{itemize}

\section{Related Work}\label{sec:related}

\textbf{Surgical instrument tracking.} Optical systems (NDI Polaris) anchor most navigation
products and EM trackers (NDI Aurora) extend tracking into the body~\cite{b1}; across six NDI
trackers, infrared held 1.6--2.0\,mm registration error and EM 2.8--3.3\,mm, metal interference
causing most EM failures~\cite{b2}. Marker-less endoscopic pose estimation~\cite{b3} degrades under occlusion --- the best SurgRIPE
entry rose from 3.0 to 12.4\,mm~\cite{b4} --- and capsule localization tracks single, slow devices
against an external array~\cite{b8}. EM sensors in form-fitting adapters have tracked through a
sterile drape in laparoscopic liver surgery~\cite{bOlthof}, the closest clinical precedent for what
we do with radio, and one that inherits the metal sensitivity of the modality. Across the
categories surveyed in~\cite{bChmarra}, each needs an external reference or a trainer-box mount;
none works with the instruments already in the surgeon's hands.

\textbf{The trocar as a kinematic constraint.} The port is a remote centre of motion (RCM): the
shaft cannot translate away from it. Robotic surgery enforces this
mechanically, by RCM linkages~\cite{bWangRCM}, or virtually, in the
controller~\cite{bFunda,bAghakhani,bLiRCM}; hand-held laparoscopy obeys the same fulcrum with only
the abdominal wall to enforce it~\cite{bGallagher}, less rigidly, but in every laparoscopic
operation. Because a shallower insertion amplifies tip displacement for the same angular
deviation~\cite{bLiRCM}, the insertion depth, not the pivot, is worth estimating. Estimation has
used the constraint only to narrow a continuous search --- pivot calibration~\cite{bYaniv}, or
laparoscopic vision searching along image lines through the insertion point~\cite{bVoros} --- never
to settle a discrete unknown. We estimate the per-pair half-wavelength integers
$n_{ij}\in\mathbb{Z}$ from the geometry, jointly with the insertion depths, and read the residual of
the same fit to tell when one is wrong.

\textbf{Range-only estimation.} A distance carries no direction: three scalars cannot place two
instruments whose antennas are free in space, and no averaging repairs an underdetermined
system~\cite{bOlson,bMartinelli}. The usual remedies add fixed anchors or accumulate measurements
until the trajectories have excited the estimate~\cite{bZhou,bTrawny}. Neither fits surgery --- the
field leaves little room for anchors, and the surgeon, not the estimator, decides the trajectories
--- so we reduce the unknowns instead: the trocar leaves each antenna one degree of freedom.

\textbf{UWB carrier-phase ranging.} UWB ToF ranging is a decimeter-accuracy positioning
technology~\cite{bJimenez}, and 802.15.4z radios expose the carrier phase, usually for
angle of arrival across an antenna array~\cite{b5}. Using that phase for distance requires
resolving the integer ambiguity~\cite{b6}; GNSS attitude work shows a known geometric constraint
can carry that burden, a fixed baseline turning the search into a constrained integer least
squares~\cite{bCompass,bGiorgi,bPark}. The trocar plays that role here. Ma \emph{et al.}
reached sub-millimeter median error with dual-frequency switching but report 10.2\,cm error with a
metal reflector 20\,cm off the line of sight~\cite{b7}; such schedules also break phase continuity
on single-PLL radios, and the surgical regime --- multi-node, fast-handheld, metal-rich, with
per-boot antenna-delay instability --- lies outside their assumptions.

\section{Methods}\label{sec:method}

\begin{figure*}[!t]
\centering
\includegraphics[width=0.9\textwidth]{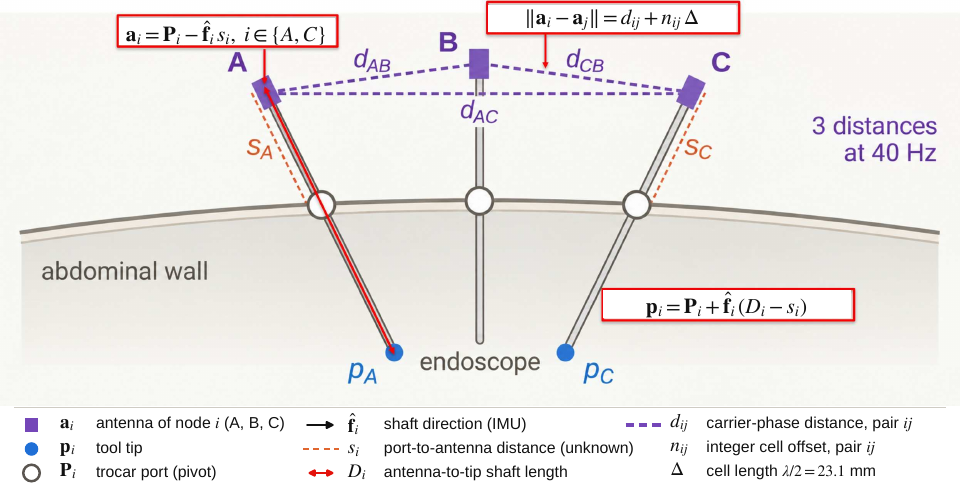}
\caption{\textbf{From three distances to two tool tips.} Each hand-held instrument pivots at
its trocar port, so its antenna and its tip lie on one line whose direction the IMU measures,
and the insertion coordinate $s_i$ is the instrument's only unknown. The three carrier-phase
distances, each known only up to an integer number of cells, therefore over-determine the two
unknowns $(s_A,s_C)$ by one, and the spare equation tests the chosen integers
(Sec.~\ref{sec:judge}). The mount offset $\mathbf R_i\mathbf o_i$ of Eq.~\eqref{eq:antenna} is
omitted from the drawing.}
\label{fig:solve}
\end{figure*}

\subsection{Platform: three UWB nodes on hand-held instruments}\label{sec:platform}
The kinematic logger is three identical nodes on one standard 5\,mm laparoscopic set-up: one
clipped to the shaft of each hand-held instrument between the handle and the trocar, and one on
the endoscope (Fig.~\ref{fig:teaser}). A node is a Qorvo DWM3001CDK development board (DW3110 UWB transceiver, nRF52833 MCU, chip
antenna, on-board LIS2DH12TR accelerometer; $75\times55$\,mm with debugger and headers, around the
$27.0\times19.1\times3.2$\,mm DWM3001C module) with a GY-BNO085 IMU supplying 100\,Hz fused
attitude, a 3.7\,V LiPo cell so that the instrument carries no cable, and a 3D-printed mount. Every node sits
outside the sterile barrier, and nothing is added to the part of the instrument that enters the
patient. Per-round ranging records and IMU samples are streamed to a host PC and time-stamped on
its clock; responder nodes piggyback their records onto the ranging exchange itself, so one host
link serves the whole set. Recordings are organized in episodes of 30--120\,s, each seeded once
by ToF. During the experiments a live display of the tracked distances was used to monitor the
session; all final analysis was computed from the recorded raw per-round logs.

\subsection{Single-channel carrier-phase ranging}\label{sec:radio}
The solve of Sec.~\ref{sec:judge} consumes a phase chain, and a chain is only as good as its
continuity: every lost or corrupted round is a chance for an integer to move. The radio layer is
therefore built around one goal, keeping the chain unbroken. Three UWB nodes --- two hand-held
instruments and the endoscope --- range pairwise on
channel~5 ($f_c{=}6489.6$\,MHz, preamble 64, turnaround 1500\,\textmu s); a controller node
interleaves double-sided two-way rounds with its two peers while a hybrid node ranges the third
pair, giving all three distances at 35--46 rounds/s per pair with no failed exchange in 125,176
rounds (Sec.~\ref{sec:res-radio}). We avoid the
usual two-channel synthetic wavelength ($\Lambda{=}100.1$\,mm): its wide-lane combination admits
near-coincident false fixes at $\pm21$ and $\pm72$\,mm, the channels cannot be sampled
simultaneously, and every channel hop re-locks the PLL and re-randomizes the carrier phase --- our
two-channel version slipped a cell every 30--60\,s. One channel quadruples the rate and removes
re-locks. A prebuffer warm-up removes the channel-5 PLL's random $0/\pi$ start-of-burst phase, crystal
trimming halves per-round phase noise (Sec.~\ref{sec:res-radio}), transmit power on the closest pair
is attenuated to avoid a saturation bias, and a 15-minute warm-up bounds residual drift.

Each round yields a ToF distance and the round-trip carrier phase $\phi\in[0,2\pi)$, in which
local-oscillator offsets cancel; a fourth, \emph{post-final} packet measures the residual
carrier-frequency term, which keeps block-mean phase coherence at 0.964 even under a forced PLL
re-lock every 2\,s (0.885 uncorrected). The phase fixes distance modulo $\Delta=\lambda/2=23.1$\,mm,
and between rounds we accumulate the wrapped increment
\begin{equation}
\delta_k=\operatorname{wrap}(\phi_k-\phi_{k-1})\,\frac{\Delta}{2\pi},\qquad
\tilde d_k=\tilde d_{k-1}+\delta_k,
\label{eq:chain}
\end{equation}
which is exact while the instrument moves less than $\Delta/2=11.5$\,mm between rounds, i.e.\ up to
400\,mm/s at 35\,Hz. A per-round quality gate (SNR and a circular-consistency test at $0.45\times$
the median SNR) rejects rounds at multipath amplitude nulls, whose phase is not the path's. ToF is not fused into $\tilde d$: the antenna delay of these development boards is uncalibrated
and its per-boot bias re-randomizes by $\pm$70--100\,mm, so ToF only seeds each episode's absolute
offset.

\subsection{Ambiguity-grid Kalman filter}\label{sec:kf}
The chained estimate can slip an integer cell when rounds are lost in motion. A scalar Kalman
filter on the inter-antenna distance extracts from each innovation $r_k$ the nearest grid component
$n_k=\operatorname{round}(r_k/\Delta)$ and \emph{absorbs} $n_k\Delta$ as a measurement artifact
unless the DWM3001CDK's on-board LIS2DH12TR accelerometer reports speed above an absorption gate
$v_{\max}$; process noise scales with the observed speed and innovations are $3\sigma$-gated. The
accelerometer only gates and scales, so its dropout degrades to pure-phase tracking. A real step of
11.55\,mm within one 0.05\,s filter cycle requires 231\,mm/s, while the accelerometer-integrated
speed exceeds a few tens of mm/s during most hand movements, so we set $v_{\max}=250$\,mm/s
(Sec.~\ref{sec:res-static} ablates 25\,mm/s). The filter cycle, not the chain, is thus the binding
speed limit, 231\,mm/s; freehand phantom manipulation reached a 95th-percentile speed of
23--42\,mm/s, a fivefold margin.

\begin{figure*}[!t]
\centering
\subfloat[Rail]{\includegraphics[height=0.39\textwidth]{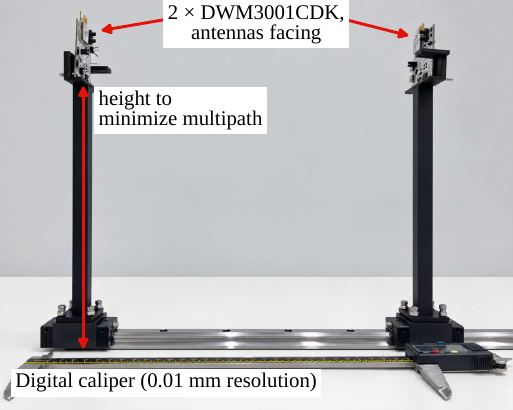}\label{fig:setup_rail}}\hspace{0.015\textwidth}%
\subfloat[Phantom]{\includegraphics[height=0.39\textwidth]{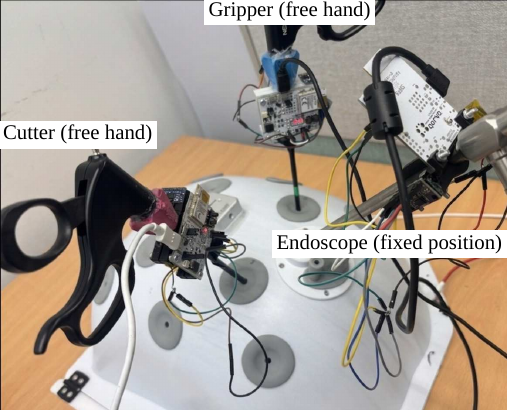}\label{fig:setup_phantom}}
\caption{\textbf{Ranging test setup.} (a)~Two DWM3001CDK nodes on 3D-printed mounts face each other on
an aluminium-profile rail with a digital caliper (0.01\,mm resolution; range 150--400\,mm); the
drape experiments use the same rail. (b)~Laparoscopic box trainer with an 11-hole trocar plate:
the endoscope node in the central port and a gripper and a cutter, each carrying a node, hand-held
through two further ports.}
\label{fig:setup}
\end{figure*}

\subsection{Trocar-constrained solve: from three distances to two tool tips}\label{sec:judge}

\textbf{Antenna on the pivot line.} Let $\mathbf P_i$ be the port of instrument $i\in\{A,C\}$ and
$\hat{\mathbf f}_i$ the unit vector along its shaft, pointing into the body, given by the on-board
IMU. Because the shaft passes through the port, its antenna lies on the line through
$\mathbf P_i$ with direction $\hat{\mathbf f}_i$, at a shaft coordinate $s_i$ behind the port,
offset laterally by the mount (Fig.~\ref{fig:solve}):
\begin{equation}
\mathbf a_i(s_i)=\mathbf P_i-\hat{\mathbf f}_i\,s_i+\mathbf R_i\,\mathbf o_i,
\label{eq:antenna}
\end{equation}
where $\mathbf R_i$ is the IMU attitude and $\mathbf o_i$ the fixed antenna offset of the mount
in the instrument frame. With the shaft length $D_i$ from antenna to working end, the tool tip is
\begin{equation}
\mathbf p_i=\mathbf P_i+\hat{\mathbf f}_i\,(D_i-s_i),
\label{eq:tip}
\end{equation}
so $D_i-s_i$ is the insertion depth and $s_i$ is the only unknown of instrument $i$. The
endoscope antenna $\mathbf a_B$ is treated as fixed and is computed once from its own port,
attitude and insertion.

\textbf{Measurement model.} A two-way carrier-phase observation on pair $ij$ yields
\begin{equation}
d_{ij}=\lVert\mathbf a_i-\mathbf a_j\rVert-n_{ij}\Delta+\varepsilon_{ij},\qquad
n_{ij}\in\mathbb Z,\ \Delta=\lambda/2,
\label{eq:meas}
\end{equation}
with $\varepsilon_{ij}$ the phase noise and $n_{ij}$ the integer number of half-wavelength cells
by which the phase chain of Eq.~\eqref{eq:chain} may have slipped since the episode's ToF seed.
Three pairs give three equations in the two continuous unknowns $\mathbf s=(s_A,s_C)$ and the
three integers $\mathbf n=(n_{AB},n_{CB},n_{AC})$.

\textbf{Mixed integer--continuous least squares.} For a fixed integer vector $\mathbf n$ the
problem is a small nonlinear least squares in $\mathbf s$,
\begin{equation}
J(\mathbf s;\mathbf n)=\sum_{ij}\big(\lVert\mathbf a_i(s_i)-\mathbf a_j(s_j)\rVert-d_{ij}-n_{ij}\Delta\big)^2,
\label{eq:ls}
\end{equation}
which we solve by Gauss--Newton from the previous frame's solution (with a grid of starts on the
first frame, since the distance equations are quadratic and admit mirrored minima). The integers
are handled by enumeration over a hypothesis set $\mathcal N$: the null hypothesis
$\mathbf n=\mathbf 0$ and single-pair corrections $n_{ij}\in\{\pm1,\dots,\pm5\}$ with the other
two pairs at zero, 31 hypotheses that cover $\pm116$\,mm, more than the per-boot ToF bias. The
solve is
\begin{equation}
(\hat{\mathbf s},\hat{\mathbf n})=\arg\min_{\mathbf s,\ \mathbf n\in\mathcal N} J(\mathbf s;\mathbf n)
\quad\text{s.t.}\quad s_A,s_C\in[60,230]\,\text{mm},
\label{eq:mils}
\end{equation}
where the depth gate rejects the degenerate solutions that the quadratic residual can otherwise
reach. Eq.~\eqref{eq:mils} is where a wrong integer becomes visible: a cell error on any single pair
cannot be absorbed by the two depths and appears in the residual.

\textbf{Residual margin and commit rule.} Let $r_0$ be the residual RMS of the null hypothesis
and $r_\star$ that of the best hypothesis. The arbiter commits a correction only when the margin
$r_0-r_\star$ exceeds 3\,mm for five consecutive solves; otherwise it stays silent and the phase
chain continues uncorrected. The margin is also the self-check the formulation provides for
free: with correct mounting offsets the three-pair residual closes at 0.4\,mm, a wrong offset
sign opens it to 36\,mm, and an uncorrected one-cell slip raises it by a few
millimetres, depending on pose (Sec.~\ref{sec:res-tip}). Once $\hat{\mathbf s}$ is fixed, Eq.~\eqref{eq:tip}
gives both tips in the endoscope frame at no extra cost; the same solve that registers the
integers is the one that localizes the tools.

\subsection{Detachable module and mounting calibration}\label{sec:integrated}
The node is applied as a clip-on module, leaving the original laparoscopic instrument
unmodified. The same module, on the same 3D-printed mount, was carried by a gripper and a cutter
in the phantom, by the endoscope, and by the surgeon's own instruments in vivo, none of which was
altered. Moving it to another instrument changes only two numbers in Eq.~\eqref{eq:antenna}, the
shaft length $D_i$ and the antenna offset $\mathbf o_i$ of the mount. In our experiments a prior
measurement found a lateral offset of 13.5\,mm and a vertical offset of 29.5\,mm from the shaft
axis, and both were entered into the model.

\section{Experiments and Results}\label{sec:results}
\subsection{Experimental setup}\label{sec:exp}

\textbf{Rail (Fig.~\ref{fig:setup_rail}).} Two DWM3001CDK nodes face each other on
3D-printed mounts on an aluminium-profile rail, at separations of 150--400\,mm, and ground truth
is obtained with a digital caliper of 0.01\,mm resolution. Both rail experiments preceded the three-node build and use a two-node link that alternates
channels~5 and~9 with a separate phase chain per channel: a two-node link has no spare equation, so
the second channel's chain serves as its internal check. The phantom and in-vivo sets run the
single-channel three-node schedule of Sec.~\ref{sec:radio}. \emph{Blind
displacement}: displacements of 25--53\,mm are set with the caliper and scored against it
($N{=}5$). \emph{Sterile drape}: at a fixed separation of 250.00\,mm, 120\,s static recordings
are taken bare, with the drape laid loosely over both nodes, and with the drape pressed onto both
chip antennas; blind displacements are then repeated with the drape pressed on (four runs; one is excluded
because no displacement was applied and no caliper value was recorded). For these displacements
the raw chains of both channels are reported as recorded, and the ToF of the same
rounds is scored against the same caliper values.

\begin{figure*}[!t]
\centering
\includegraphics[width=\textwidth]{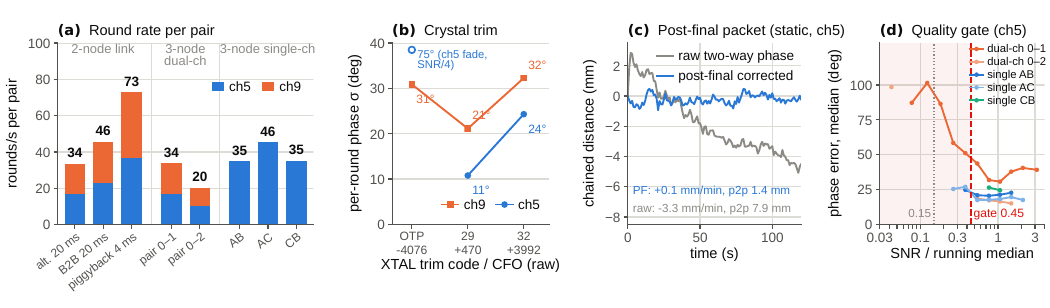}
\caption{\textbf{Four mechanisms that keep the carrier phase continuous from round to round.}
The distance estimate accumulates phase increments, so a lost or corrupted round is a chance for the
integer to move; each panel isolates one defence. (a)~Ranging rate per pair: the more rounds a chain
receives, the less the instrument can move between them. The packet path raised the two-node link
from 34 to 73 rounds/s, and the single-channel three-node schedule delivers 35--46 rounds/s per pair,
where the dual-channel schedule left each channel's own chain only 10--17. (b)~Per-round phase noise
against the crystal trim that sets the carrier-frequency offset; the hollow point is confounded by a
channel-5 fade. (c)~A static link should trace a flat line: without the post-final packet the chain
drifts, with it the drift disappears. (d)~Median phase error of the rounds in each SNR bin, relative
to the running median SNR; the rounds left of the gate carry the largest errors and are discarded.}
\label{fig:radio}
\end{figure*}

\begin{figure*}[!t]
\centering
\includegraphics[width=\textwidth]{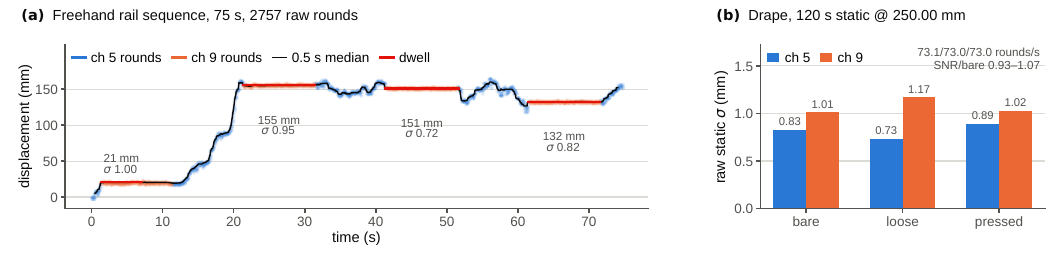}
\caption{\textbf{Rail experiments, two-node link alternating channels 5 and 9.} (a)~A 75\,s freehand
sequence, 2,757 raw rounds; red bars mark the detected dwells. (b)~Raw static $\sigma$ at 250.00\,mm,
120\,s each, bare, with the drape laid loosely over both nodes and pressed onto both antennas.
(c)~Blind displacements with the drape pressed on: both phase chains as recorded, the ToF of the
same rounds (gray) and the caliper value (dotted). (d)~Endpoint error of each chain and of ToF on a
symmetric-log axis; the red lines mark one cell.}
\label{fig:rail}
\end{figure*}

\textbf{Phantom (Fig.~\ref{fig:setup_phantom}).} A laparoscopic box trainer with an 11-hole trocar
plate holds the endoscope node in the central port and a gripper and a cutter, each carrying a node,
hand-held through two further ports. \emph{Channel choice}: static sessions of the instrument pair on channel~9 and on
channel~5. \emph{Static and freehand manipulation}: 39 usable episodes in one session, 22 freehand
(median 28\,s) and 17 stationary in the trocars (40--57\,s), two shorter than 10\,s excluded;
distances and per-pair ToF are logged at 10\,Hz, and an ablation repeats the freehand protocol at
$v_{\max}=25$\,mm/s (16 episodes). An integer-slip candidate is a 0.2--0.6\,s window in which one
pair changes by $k\Delta\pm4$\,mm ($k{=}1$--3) and the other two by less than 3\,mm; that pair's ToF,
independent of the phase grid, labels it \emph{real motion}, \emph{slip} or \emph{undecided}, and
motion exposure counts the time a pair changes by at least 2\,mm within 3\,s. \emph{Tool-tip
localization}: with the instruments in known holes we score the closure of the solve of
Sec.~\ref{sec:judge} and the repeatability of a parked tip while the other instrument moves.

\textbf{In vivo (Fig.~\ref{fig:invivo}).} The logger was carried into a live rabbit appendectomy
in the operating room with the same settings as in the phantom; the procedure was approved by the
institutional animal care and use committee (details withheld for review).

\textbf{Protocol and baselines.} A 15-minute warm-up precedes every precision measurement. The
baselines are ToF from the same radio and the same rounds, the raw phase chain of
Eq.~\eqref{eq:chain}, and a rule-based supervisory stack without the Kalman filter.

\subsection{Radio layer: keeping the phase chain continuous}\label{sec:res-radio}
Packet-path optimization raised the two-node rate from 33.5 to 73.1 rounds/s, and the
three-node single-channel schedule sustained 35.1, 45.5 and 35.2 rounds/s on $AB$, $AC$ and $CB$
with no failed exchange (Fig.~\ref{fig:radio}). Crystal trimming halved the per-round phase noise
(24.3$^\circ\to$10.8$^\circ$ on channel~5), the post-final correction suppressed static drift from
$-3.3$ to $+0.1$\,mm/min, and the quality gate discarded the low-SNR rounds, whose phase error was
twice as large; fewer than 0.2\% of rounds fell below it.

\subsection{Rail: continuity and the sterile drape}\label{sec:res-rail}
\textbf{Continuity.} On a 75\,s freehand sequence the raw chain followed every transition and held
four dwells at 21, 155, 151 and 132\,mm with $\sigma=0.72$--1.00\,mm (Fig.~\ref{fig:rail}a).
On the bare rail, five blind displacements of 25--53\,mm gave signed errors of
$+1.8/+4.6/-0.1/-2.1/+5.3$\,mm against the caliper, a mean absolute error of 2.8\,mm with zero grid
slips; with a 0.01\,mm reference this error belongs to the fixture and the radio.

\textbf{The drape leaves the link unchanged.} Bare, loosely draped and with the drape pressed onto
both chip antennas, the round rate was 73.1, 73.0 and 73.0 rounds/s, the first-path SNR of each board
stayed within 0.93--1.07 of its bare value, and the raw static $\sigma$ was 0.83/0.73/0.89\,mm on
channel~5 and 1.01/1.17/1.02\,mm on channel~9 (Fig.~\ref{fig:rail}b). The spread follows the channel,
not drape contact. Single-shot ToF on the same rounds scattered by 17--20\,mm on channel~5 and
28--148\,mm on channel~9. With the drape pressed on, the two chains gave
$-27.21$ and $-24.10$\,mm for $-28.10$\,mm (run~1),
$+28.82$\,mm on channel~9 for $+28.10$\,mm (run~2), and $+49.82$\,mm on channel~5 for $+50.00$\,mm
(run~4); taking in each run the chain that held its registration, the three
displacements were tracked at 0.81\,mm mean absolute error. ToF of the same rounds erred by
6.7--149\,mm (Fig.~\ref{fig:rail}c,~d).

\begin{figure}[!t]
\centering
\includegraphics[width=\columnwidth]{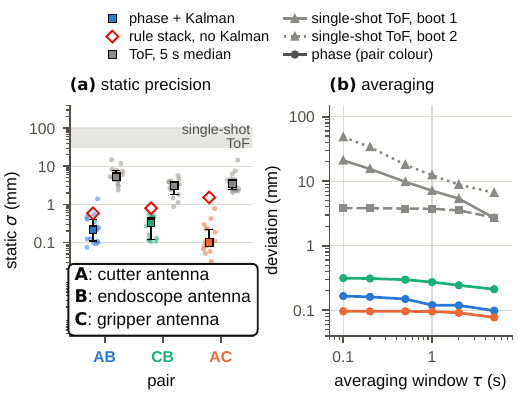}
\caption{\textbf{Static precision in the phantom.} (a)~Static $\sigma$ of the tracked distance in
each stationary episode, sampled at 0.5\,Hz, on a log axis: carrier phase with the Kalman filter
(colored dots; squares, median and bootstrap 95\% interval), the rule stack without the filter
(diamonds; one 240\,s live capture), and ToF of the same rounds (gray; already a 5\,s median, hence a
lower bound for ToF). The gray band marks single-shot ToF on a static bench pair. (b)~Deviation
against the averaging window: the phase sits at its floor from 0.1\,s, while ToF, single-shot or
5\,s median, stays an order of magnitude above it at every window.}
\label{fig:static}
\end{figure}

\subsection{Phantom: static precision, ToF and grid integrity}\label{sec:res-static}
Metal instruments and oblique antennas make the multipath harsher here than on the rail.

\textbf{Channel choice.} In the phantom, static sessions of the instrument pair made 37 and 17
integer-cell jumps per session on channel~9 against 1 on channel~5, whose 23\% larger half-cell
margin motivated the single-channel design.

\textbf{Precision.} With the instruments resting in their trocars, the tracked distance held a median $\sigma$ of 0.22,
0.33 and 0.10\,mm on $AB$, $CB$ and $AC$ (bootstrap 95\% intervals 0.11--0.41, 0.12--0.45 and
0.08--0.22\,mm) over the 15, 14 and 16 stationary episodes in which each pair stayed still
(Fig.~\ref{fig:static}).

\textbf{Why ToF only seeds the offset.} Across two power cycles of the same static bench pair, the ToF median moved by $+35.2$\,mm, the single-shot $\sigma$ grew from 31
to 105\,mm, and the second boot turned bimodal. A bias that re-draws at every
boot cannot be averaged away, so ToF sets only where an episode starts.

\textbf{Grid integrity.} Over the 22 freehand episodes (30.7 pair-minutes of motion) the window
detector raised 31 candidates ($AB$ 6, $AC$ 21, $CB$ 4). Twenty carried a ToF witness: three were
real motion, and none fell in the slip band where the phase steps by a cell while ToF stays within
8\,mm, so no slip was confirmed and the one-sided 95\% bound is 0.29 slips per pair-minute, or 1.2,
2.7 and 0.8 when every undecided candidate is counted as a slip. The ablation shows that the
absorption gate is what buys this: at $v_{\max}=25$\,mm/s, over a comparable 31.4 pair-minutes, the
same detector and witness confirmed six slips, three on $AB$ and three on $AC$, an imbalance with an
exact conditional probability of 0.015.

\begin{figure}[!t]
\centering
\includegraphics[width=\columnwidth]{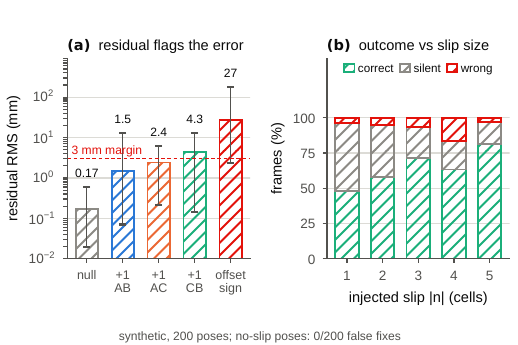}
\caption{\textbf{Trocar-constrained solve} (synthetic: ranges drawn with 0.5\,mm noise on the
measured phantom geometry, 200 poses per point). (a)~Residual RMS of the null solve, of an
uncorrected one-cell slip on each pair and of a wrong mounting-offset sign; bars are medians with
5th--95th percentile whiskers, and the red line is the 3\,mm commit margin. (b)~Outcome of the solve
against injected slip size: correct correction, silence, or a wrong cell.}
\label{fig:tip}
\end{figure}
\subsection{Trocar-constrained solve}\label{sec:res-tip}
\begin{figure}[!tb]
\centering
\includegraphics[width=0.90\columnwidth]{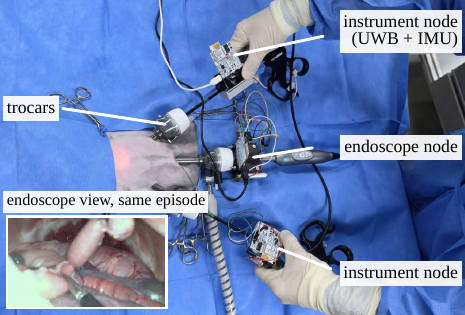}
\caption{\textbf{In-vivo deployment:} appendectomy on a live rabbit, a node on each hand-held
instrument and a third on the endoscope, all outside the trocars. Inset: the endoscope view of the
same episode.}
\label{fig:invivo}
\end{figure}
\textbf{On the phantom.} With correct mounting offsets the solve closed at a three-pair residual RMS
of 0.4\,mm (36\,mm for the wrong offset sign), and a stationary
instrument's solved tip held within a median 1.8\,mm while the other was manipulated.

\textbf{The spare equation is a self-check.} A wrong assumption cannot be absorbed by the two
depths and has to show up in the residual. With everything correct the solve closed at a median residual RMS of 0.17\,mm, the level
of the range noise itself. An uncorrected one-cell slip raised it to 1.5, 2.4 and 4.3\,mm on $AB$,
$AC$ and $CB$, and to about 13\,mm in the worst poses: visible, but often below the 3\,mm commit
margin, which is why the arbiter stays silent on many one-cell slips. A wrong mounting-offset sign
raised it to 26.9\,mm, so a setup error of that kind cannot pass unnoticed (Fig.~\ref{fig:tip}a).
These numbers measure how strongly a wrong assumption shows in the residual, not the accuracy of the
estimate.

\textbf{Integer registration.} Over 200 slip-free poses the $\pm5$-cell hypothesis set committed no
false fix. Over 200 injected slips the solve committed the correct correction in 126 (63\%), stayed
silent in 61 and picked a wrong cell in 13, so 82\% of its failures were silence
(Fig.~\ref{fig:tip}b).

\textbf{Tip error.} With exact attitude, 0.5\,mm range noise gives a median tip error of 0.92\,mm
(95th percentile 5.4\,mm). An IMU attitude error of 1$^\circ$ and 2$^\circ$ raises the median to 3.6
and 7.4\,mm, because the error enters through both the lever arm and the solved
depth; attitude, not range, sets tip accuracy, which is where the next gain lies.

\subsection{In-vivo deployment}\label{sec:res-invivo}
During live rabbit appendectomy the kinematic logger was mounted on the surgeon's hand-held
instruments and operated through the trocars with the same mounts and logging chain as in the
phantom, nothing added to the part that enters the animal (Fig.~\ref{fig:invivo}). It logged 11
episodes, 133\,s of three-pair ranging and 6.0 pair-minutes of motion exposure, with the working ends
inside the animal.

\section{Discussion and Limitations}\label{sec:limits}
\textbf{What the evidence supports.} Two things carry the method. Distance is the one quantity a
sterile drape leaves intact, and carrier phase makes it precise enough to be a geometric
measurement rather than a proximity cue. The trocar then supplies the direction that distance
lacks, and combining the two is what makes a tool-tip estimate possible.

\textbf{In-vivo tip localization is not yet validated.} The in-vivo recording needs more
episodes, and one instrument's IMU often produced incomplete attitude, so those records hold
distances rather than solved tips. Converting a solved depth into a tip also needs the shaft
direction: at the 1--2$^\circ$ of a fused IMU the median tip error is 3.6--7.4\,mm, so the system
delivers sub-millimeter distances and millimeter-level tips, and better attitude buys more than
better ranging.

\textbf{Ranging is relative, and the integer is bounded rather than proven.} Each episode is seeded
by ToF, whose bias re-draws at every power cycle, so absolute tip accuracy against surveyed
fiducials was not measured. No ToF-confirmed slip occurred in the freehand phantom sessions, but
when ToF wandered during manipulation some candidates could not be decided.

\textbf{Practical limits.} The pivot is not perfectly fixed --- the abdominal wall is
viscoelastic~\cite{bKirilova} and control work already treats the RCM as time-varying~\cite{bKastritsi}
--- so a port that shifts during manipulation biases the solved depth. The nodes are development
boards on 3D-printed mounts and need sterilizable packaging before clinical use; the endoscope node is treated as fixed, so a moving
endoscope would add a third insertion depth, which its own IMU and trocar supply in the same form
as the instruments; and the rail validation is limited by its
fixture, which measures displacement against a caliper rather than absolute distance.

\section{Conclusion}\label{sec:conclusion}
We presented a kinematic logger that clips three ultra-wideband nodes onto a standard laparoscopic
set-up and estimates tool geometry from distances alone. Carrier phase on a single channel holds the
three inter-antenna distances to $\sigma=0.10$--$0.33$\,mm among metal instruments and survives a
sterile drape pressed onto the antennas, while the trocar reduces two free instruments to two
insertion depths with one equation to spare. The same solve returns both tool tips in the endoscope frame, and
the logger recorded a live rabbit appendectomy with nothing added inside the animal.

\section*{ACKNOWLEDGMENT}
Acknowledgements of the surgeons and funding sources will be added after ICRA 2027
acceptance.

\end{document}